%% file: Input_Loss_Landscapes__ILL__Reveal_Residual_Memorization_in_Large_Language_Models_After_Unlearning.tex
\documentclass[10pt, a4paper]{article}
\usepackage{lrec2026}

\usepackage{multibib}
\usepackage{graphicx}
\usepackage{tabularx}
\usepackage{soul}
\usepackage{url}
\usepackage{amsmath}
\usepackage{amsfonts}
\usepackage{algorithm}
\usepackage{algorithmic}
\usepackage{tikz}
\usetikzlibrary{positioning, calc}
\usepackage{float}
\usepackage{array}
\usepackage{makecell}
\usepackage{wrapfig}

\usepackage{booktabs} 
\usepackage{wrapfig}
\usepackage{floatflt}
\usepackage{multirow}
\usepackage{microtype}
\usepackage{makecell}
\usepackage{subcaption}

\usepackage[table]{xcolor}

\title{
    REMIND: Input Loss Landscapes Reveal Residual Memorization in Post-Unlearning LLMs \thanks{This is a pre-print version under review.}
    }

\name{Liran Cohen, Yaniv Nemcovesky, Avi Mendelson }
\address{   Technion - Israel Institute of Technology \\
            {liranc6@campus.technion.ac.il, 
            yanemcovsky@campus.technion.ac.il,
            avi.mendelson@technion.ac.il}
        }

\abstract{
Machine unlearning aims to remove the influence of specific training data from a model without requiring full retraining.
This capability is crucial for ensuring privacy, safety, and regulatory compliance. Therefore, verifying whether a model has truly forgotten target data is essential for maintaining reliability and trustworthiness.
However, existing evaluation methods often assess forgetting at the level of individual inputs. This approach may overlook residual influence present in semantically similar examples. Such influence can compromise privacy and lead to indirect information leakage.
We propose \textbf{REMIND (Residual Memorization In Neighborhood Dynamics)}, a novel evaluation method aiming to detect the subtle remaining influence of unlearned data and classify whether the data has been effectively forgotten. \textsc{REMIND} analyzes the model's loss over small input variations and reveals patterns unnoticed by single-point evaluations.
We show that unlearned data yield flatter, less steep loss landscapes, while retained or unrelated data exhibit sharper, more volatile patterns.
\textsc{REMIND} requires only query-based access, outperforms existing methods under similar constraints, and demonstrates robustness across different models, datasets, and paraphrased inputs, making it practical for real-world deployment. By providing a more sensitive and interpretable measure of unlearning effectiveness, \textsc{REMIND} provides a reliable framework to assess unlearning in language models. As a result, \textsc{REMIND} offers a novel perspective on memorization and unlearning.
We provide a complete \textsc{REMIND} implementation for easy unlearning evaluation: \url{https://anonymous.4open.science/r/Input-Loss-Landscapes-ILL-Reveal-Residual-Memorization-in-Post-Unlearning-LLMs-3FC0/}
\newline\newline
}
\begin{document}

\maketitleabstract

\section{Introduction}

\begin{figure*}[htb]
\centering
\begin{subfigure}[t]{0.3\textwidth}
    \centering
    \caption{Forgotten example}
    \includegraphics[width=\textwidth]{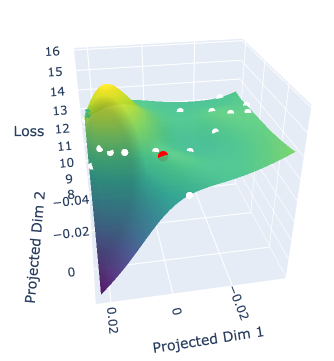}
    \label{fig:ill_forgotten}
\end{subfigure}%
\hfill
\begin{subfigure}[t]{0.3\textwidth}
    \centering
    \caption{Holdout example}
    \includegraphics[width=\textwidth]{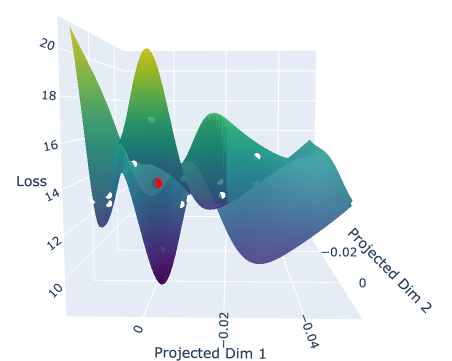}
    \label{fig:ill_holdout}
\end{subfigure}%
\hfill
\begin{subfigure}[t]{0.3\textwidth}
    \centering
    \caption{Retained example}
    \includegraphics[width=\textwidth]{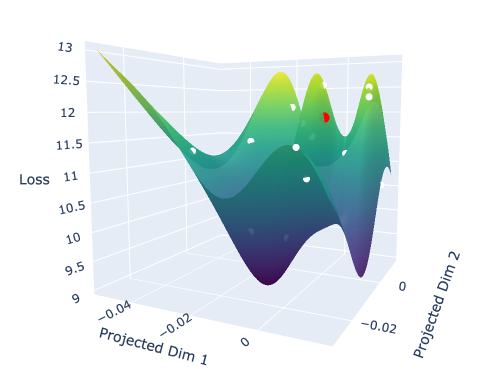}
    \label{fig:ill_retained}
\end{subfigure}
\vspace{-2ex}
\caption{Input Loss Landscapes (ILL) for forgotten, holdout, and retained examples. Showing the input target (red) and its perturbed neighbors (white). Loss is visualized on a 2D embedding space via dimensionality reduction.}
\label{fig:ill_examples}
\end{figure*}
Machine unlearning is the task of removing the influence of specific data from a trained model. It has become increasingly important as LLMs are deployed at scale. These models are often trained on extensive datasets, which may include sensitive or copyrighted content, raising legal and ethical concerns about privacy and data ownership \cite{machine_unlearning_survey}. Regulations such as the General Data Protection Regulation (GDPR) \cite{gdpr} mandate the "right to be forgotten", requiring the removal of personal data upon request. While full model retraining can satisfy such requirements, it is typically infeasible for large models, leading to growing interest in unlearning methods that offer targeted data removal. However, a key challenge remains: how can we evaluate whether unlearning has truly removed the unlearned data influence from the model's behavior?

\subsection{Current Approaches and Their Limitations}
Black-box evaluation methods observe only a model's inputs and outputs. They focus on single-input sentences and measure behavior or loss at individual points. Such approaches assume that forgetting is a pointwise phenomenon. Such a perspective can be overly confident and overlook forgetting patterns in the semantic neighborhood of an input. As a result, those methods might fail to detect when the model still reflects forgotten data.

In contrast to black-box evaluation methods, white-box methods leverage internal signals such as parameters, gradients, or activations. These signals can reveal different aspects of what the model retains or forgets. However, the access they require is often impractical, and their computational cost can be prohibitive in real-world settings.

\subsection{Residual Memorization}

Residual memorization refers to lingering traces of forgotten data in a model, which remain after that data has undergone the unlearning process \citep{hsu2025are}. This residual influence often manifests subtly, through the model's behavior on inputs that are semantically similar to the unlearned data, even if the original sample is no longer explicitly recalled. Such behavior undermines the goals of unlearning by allowing indirect leakage or influence of unlearned data to persist.

\subsection{The Input Loss Landscape as a Diagnostic Tool}

The Input Loss Landscape (ILL) captures how a model's loss responds to small perturbations in the input, providing an interpretable measure of sensitivity and robustness. We demonstrate that the ILL reveals clear evidence of residual memorization signals in machine unlearning. Specifically, as illustrated in Figure~\ref{fig:ill_examples}, forgotten data corresponds to flatter and shallower regions, while retained data shows sharper and more volatile patterns. The ILL's reliance solely on inputs and outputs without access to model weights makes it compatible with black box approaches. It also enhances explainability, as it reflects semantically meaningful input variations.

\subsection{Introducing REMIND: Neighborhood-Based Evaluation of Unlearning}

We propose \textbf{Residual Memorization In Neighborhood Dynamics (REMIND)}, a novel black-box evaluation methodology aiming to accurately identify unlearned data points by probing the Input Loss Landscape (ILL) around each sentence through coherent neighbors generated by embedding-proximity perturbations. To the best of our knowledge, prior work on unlearning evaluation in LLMs has not leveraged neighboring sentences of a target to detect residual memorization. This methodology provides a valuable resource for the NLP community to systematically assess unlearning effectiveness across model architectures and datasets, supporting evaluations in production settings where only black-box access is available.

\subsection{Contributions}

Our main contributions are: 
\begin{itemize}
    \item We reveal a key phenomenon of memorization and unlearning, where they create geometric structures in the ILL.
    
    \item We demonstrate that unlearning requires neighborhood-level evaluation, which captures critical subtle signals overlooked by pointwise metrics.
    
    \item We introduce \textsc{REMIND}, a practical evaluation method that leverages these hidden unlearning signatures to accurately identify unlearned points.

    \item Our results indicate that \textsc{REMIND} outperforms existing black-box methods in uncovering unlearning points, and is robust across datasets, architectures, and paraphrased variants.
    
\end{itemize}

\section{Related Work}

Evaluating machine unlearning is a multifaceted challenge. A useful taxonomy suggests three key dimensions, including unlearning effectiveness, unlearning efficiency, and model utility \citep{unlearningTaxonomy}. This work focuses on unlearning effectiveness, which examines the thoroughness of data erasure and the degree to which a model has genuinely forgotten specific information. It addresses the questions of what “effective forgetting” looks like in practice and how we should measure it in ways that capture subtle residual memorization.
\\ \\
Benchmarks like TOFU \citep{tofu}, MUSE \citep{muse}, WMDP \citep{wmdp}, and RWKU \citep{rwku} use membership inference attack (MIA) based metrics such as QA scores, Zlib compression \citep{zlib}, ROUGE-L F1 \citep{rouge_lin_2004}, and MIN-K\%++ \citep{minkpp} to evaluate forgetting and memorization. 

However, these methods are known to exhibit various issues \citep{critical_reexamination}. For instance, they often rely on model responses to exact prompts. Specifically, Zlib may confuse memorization with generalization, ROUGE-L F1, which focuses on lexical overlap, and QA-based methods are sensitive to prompt design. MIN-K\%++ better targets memorized content, but can overestimate privacy \citep{critical_reexamination}.
To achieve more precise and fine-grained assessments, per-sample MIAs, such as U-LiRA \citep{inexact_unlearning}, are now utilized in frameworks like RWKU and OpenUnlearning \citep{openunlearning}. However, these methods still rely on model responses to the exact input sentence.

White-box evaluation approaches as gradient tracing and internal probes \citep{intrinsic_parametric_knowledge_traces} provide deeper insights but require full model access and often depend on specific architectures or unlearning algorithms.

ILL analysis has been used in computer vision to study model robustness, overfitting, and generalization \citep{ILL_input_gradient_regularization, ILL_robust_overfitting, ILL_adversarial_weight_perturbation}. These works show that the ILL curvature can reveal vulnerabilities and signal weak generalization or memorization. In natural language processing (NLP), \citet{ILL_sharpness_adversarial_samples} observed that adversarial examples tend to lie in sharper ILL regions. However, ILL-based analyses remain underexplored in NLP and, to our knowledge, have not been applied to evaluate unlearning or residual memorization in large language models (LLMs).

Some MIAs have explored input-neighbors to assess membership \citep{neighbourhood_comparison, NoisyNeighbors, SPV_MIA, SMIA, xu2024targeted}.
However, these approaches typically rely on summary statistics such as maximum or average neighbor loss, and often require white-box access or shadow model training. As a result, they miss important patterns in the local ILL. Our work addresses this gap by analyzing the ILL full geometry surrounding the target, presenting the first approach that examines neighboring sentences to detect residual memorization.\\
Perturbation-based methods analyze and attack neural networks by altering input features and measuring how these changes affect model outputs. In natural language processing (NLP), this can involve inserting, removing, or replacing tokens within an input sentence to measure the impact on the model's confidence or prediction. Techniques like the ``Fast Gradient Sign Method (FGSM)'' perturb an input in the direction of its gradient to achieve this effect \citep{adversarial_machine_learning}.
On the white-box side, \cite{NoisyNeighbors} adds noise to the original sentence embeddings. On the black-box side, \cite{SPV_MIA} replaces words in the original sentence with masked language model (MLM) predictions. Other methods include token swaps \citep{tokens_swapping} or prefix addition \citep{recall} or suffix addition \citep{suffix_addition}. However, many of these methods produce unnatural or hard-to-interpret variations. Moreover, they lack control over the distance from the original input, require an external model such as a masked language model, and do not guarantee semantic similarity. Our method uses embedding-guided perturbations to generate coherent, similar neighbors at controlled distances, enabling faithful and resource-efficient exploration of the local ILL with minimal external dependencies.\\



\section{Method}

\begin{figure}[htb]
\centering
\begin{tikzpicture}[node distance=1.5cm, auto, >=stealth, scale=0.6, every node/.style={transform shape}]
    \node[draw, rounded corners, fill=blue!10, minimum width=2.5cm, minimum height=1cm] (input) {Input sequence $\mathbf{x}$};
    \node[draw, rounded corners, fill=green!10, below of=input, minimum width=3.2cm, minimum height=1cm] (perturb) {Embedding-proximity perturbations};
    \node[draw, rounded corners, fill=yellow!10, below of=perturb, minimum width=2.8cm, minimum height=1cm] (loss) {Model loss computation};
    \node[draw, rounded corners, fill=orange!10, below of=loss, minimum width=2.8cm, minimum height=1cm] (features) {ILL feature extraction};
    \node[draw, rounded corners, fill=red!10, below of=features, minimum width=2.8cm, minimum height=1cm] (classifier) {Classifier: retained / forgotten / holdout};
    \draw[->, thick] (input) -- (perturb);
    \draw[->, thick] (perturb) -- (loss);
    \draw[->, thick] (loss) -- (features);
    \draw[->, thick] (features) -- (classifier);
\end{tikzpicture}
\caption{Methodology overview: \textsc{REMIND} evaluation pipeline from input to diagnostic output.}
\label{fig:methodology_overview}
\end{figure}
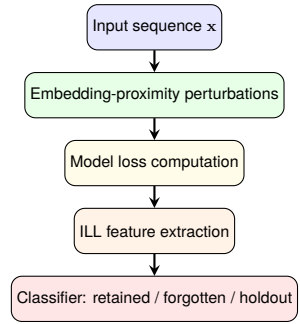

To uncover evidence of unlearning, we analyze the model's loss landscape structure around the original input and its semantically similar variants.

Our evaluation methodology comprises two key components: the \textbf{perturbation environment}, a set of semantically similar variants of the original input generated via embedding-guided perturbations; and the corresponding \textbf{Input Loss Landscape (ILL)}, defined by loss values over these perturbations as a local geometric structure.

This approach captures subtle patterns of residual memorization that are invisible to point-based metrics, particularly in black-box settings.


We extract a \textbf{feature vector} from the ILL, incorporating both raw loss values and additional structural properties:

\begin{align*}
\mathbf{feature\_vector} = [
&\ell_{\mathrm{orig}} \quad , \quad \mu_{\mathrm{neigh}} \quad , \quad \ell_{\mathrm{max}}, \\[0.5ex]
&\ell_{\mathrm{min}} \quad , \quad \sigma_{\mathrm{neigh}} \quad , \quad \sigma^2_{\mathrm{neigh}}, \\[0.5ex]
&\Delta \mu \quad , \quad \Delta_{\mathrm{max}} \quad , \quad \Delta_{\mathrm{min}}, \\[0.5ex]
&\sigma^2_{\Delta} \quad , \quad \mu_{\nabla} \quad , \quad \nabla_{\mathrm{max}}, \\[0.5ex]
&\sigma^2_{\nabla} \quad , \quad \nu_{\mathrm{neigh}} 
]
\end{align*}

Where:
 $\ell_{\mathrm{orig}}$: original loss ; 
 $\mu_{\mathrm{neigh}}$, $\ell_{\mathrm{max}}$, $\ell_{\mathrm{min}}$: mean, max, and min neighbor losses ;
 $\sigma_{\mathrm{neigh}}$, $\sigma^2_{\mathrm{neigh}}$: standard deviation and variance of neighbor losses ;
 $\Delta \mu$, $\Delta_{\mathrm{max}}$, $\Delta_{\mathrm{min}}$: mean, max, and min loss deltas relative to neighbors ;
 $\sigma^2_{\Delta}$: variance of loss deltas ;
 $\mu_{\nabla}$, $\nabla_{\mathrm{max}}$: mean and max gradients with respect to embeddings ;
 $\sigma^2_{\nabla}$: variance of gradients ;
 $\nu_{\mathrm{neigh}}$: neighborhood loss volatility .

Note that gradient features are computed using an external text encoder.
\\ \\
This feature vector is passed to a lightweight classifier trained to distinguish among three classes: \texttt{Retained} (the model has clearly not forgotten the input), \texttt{Forgotten} (the model has unlearned the input), and \texttt{Holdout} (an unrelated sentence, used as control).



\subsection{Embedding-Proximity Perturbations}

Formally, let $V$ be a token vocabulary, let $\mathbf{x_{1:n}} = \{x_i\}_{i=1}^n \in V^n$ be a textual input sequence, let $R_p(x_i, x_i') \in \{x_i, x_i'\}$ be a random replacement function with probability $p$, let $U(1,m)$ be a discrete uniform distribution over some $m \in \mathbb{N}$, let $x_i^{(j)} \in V$ be the $j$'th nearest neighbor of $x_i$ based on the corresponding embeddings' cosine similarity, denoted as $\mathrm{COSIM}(x_i, x_i')$, let $K$ denote the number of sampled perturbed variants. Then the neighborhood set is defined as:
\begin{align}
S^K = \{ \{R_p(x_i, x_i^{U(1,m)})\}_{i=1}^n \}_{\alpha=1}^K
\end{align}
Here, $S^K$ is a set of meaning-preserved variants, preserving the original input's meaning while introducing limited semantic variation. We use $S^K$ to compute first-order and second-order statistics over the evaluated model loss, including mean, variance, and volatility, which form the ILL feature vector components.

The embedding-proximity perturbation procedure is detailed in Algorithm \ref{alg:embedding_perturb}.

\begin{algorithm}[ht]
\caption{Embedding-Proximity Perturbation}
\label{alg:embedding_perturb}
\begin{algorithmic}[1]
    \REQUIRE Input sequence $\mathbf{x} = \{x_1, \dots, x_n\}$, perturbation probability $p$, number of neighbors $m$, number of perturbations $K$
    \ENSURE Set of perturbed variants $\{\tilde{\mathbf{x}}_j\}_{j=1}^K$
    \FOR{$k = 1$ to $K$}
        \FOR{$i = 1$ to $n$}
            \STATE Sample $j \sim U(1, m)$
            \STATE Let $x_i^{(j)}$ be the $j$th nearest neighbor of $x_i$ by cosine similarity
            \STATE Replace $x_i$ with $x_i^{(j)}$ with probability $p$.
        \ENDFOR
        \STATE Store perturbed sequence $\tilde{\mathbf{x}}_k = \{x_1', \dots, x_n'\}$
    \ENDFOR
    \RETURN $\{\tilde{\mathbf{x}}_j\}_{j=1}^K$
\end{algorithmic}
\end{algorithm}

\subsection{Diagnostic Pipeline}

The extracted ILL feature vector is fed into a lightweight classifier (e.g., logistic regression, random forest) trained to discriminate between retained, forgotten, and holdout data. This facilitates the detection of forgetting behaviors across both original inputs and semantically similar variants, enhancing unlearning evaluation. The pipeline remains modular and interpretable, relying solely on model loss outputs.

The full \textsc{REMIND} evaluation procedure is summarized in Algorithm \ref{alg:REMIND_alg}.

\begin{algorithm}[H]
\caption{REMIND Evaluation Procedure}
\label{alg:REMIND_alg}
\begin{algorithmic}[1]
    \REQUIRE Model $\mathcal{M}$, input $\mathbf{x}$, neighborhood size $m$, number of perturbations $K$
    \ENSURE Classification of input: Retained / Forgotten / Holdout
    \STATE Generate $K$ perturbed variants $\{\tilde{\mathbf{x}}_j\}$ using embedding-proximity with $m$ neighbors
    \STATE Compute loss $\ell(\tilde{\mathbf{x}}_j)$ for all variants
    \STATE Extract ILL features from loss distribution
    \STATE Classify input using trained classifier
\end{algorithmic}
\end{algorithm}

\section{Theoretical Motivation}

\textsc{REMIND} essentially assumes that models exhibit local smoothness around input samples. In practice, models are trained with objectives and regularization encouraging such behavior. As a result, memorization effects extend beyond individual training points into their surrounding neighborhoods. Thereby, variations of memorized sequences exhibit similar loss patterns, reflected in the ILL geometry. Distinct geometries, such as \textit{shallow loss landscapes}, where loss values remain relatively flat (low variance, low gradient magnitude, low volatility), or \textit{sharp valleys}, where loss changes abruptly (high variance, strong gradients, high volatility), can indicate memorization even when the exact sequence is not queried, giving an advantage over point-based methods. This motivation is illustrated in Figure~\ref{fig:ill_examples}, which shows how a noisy query near a memorized point reveals informative neighborhood structure due to the model's smooth generalization.

\section{Experiments}

Our evaluation focuses on the recognition of forgotten, retained, and holdout samples, and seeks to address three key research questions:

\begin{itemize}
    \item \textbf{Q1:} \textit{Can the ILL structure uncover otherwise unnoticed information about the unlearning process?}
    \item \textbf{Q2:} \textit{How does REMIND compare to pointwise and black-box baselines in uncovering forgetting patterns?}
    \item \textbf{Q3:} \textit{Do ILL-based signals remain consistently informative across various samples, architectures, and unlearning techniques?}
\end{itemize}

\subsection{Experimental Setup}

\begin{itemize}
    \item We compare \textsc{REMIND} to existing black-box unlearning evaluation metrics: naive loss-based, ROUGE-L \citep{rouge_lin_2004}, Zlib Compression \citep{zlib}, MIN-K\% \citep{minkpp}, and MIN-K\%++ \citep{minkpp}, and simplified SPV-MIA with max and mean options \citep{SPV_MIA}, all adapted to our unlearning evaluation context.

    \item We evaluate across three benchmarks: MUSE \citep{muse}, TOFU \citep{tofu}, and WMDP \citep{wmdp}, with corresponding models and unlearning methods including elm \citep{ELM}, tar \citep{TAR}, pbj, rmu-lat \citep{RMU}, rmu \citep{RMU},
    nd simNPO \citep{simNPO}.
    
    \item We evaluate across the following instruction-tuned language models: LLaMA-3-8B-Instruct \citep{grattafiori2024llama3}, LLaMA-2-7B-Chat \citep{touvron2023llama2}, and Zephyr-7B-Beta \citep{tunstall2023zephyr}.

    \item We use the GPT-2 \citep{GPT2} tokenizer in our experiments. Token-proximity perturbations require a token-to-embedding mapping, typically provided by the tokenizer. Using the same tokenizer offers practical, efficient, and fair evaluation across experiments. It removes the need for target model information and reduces computational cost. Strong results can imply robustness even with a potentially suboptimal tokenizer. A theoretical justification is that effective tokenizers preserve semantic similarity, keeping nearby embeddings meaningful and producing similar neighbors across tokenizers, even if the order may vary.



    \item Our primary evaluation metric is ROC-AUC, assessing the ability to distinguish between: \texttt{Retained vs Forgotten}, \texttt{Forgotten vs Holdout}, and all three classes jointly.
    Additionally, we report: \texttt{ROC-AUC at 1\% FPR (for low-FPR detection quality)}, \texttt{F1-score}, and \texttt{Accuracy} for completeness.

    \item We assume access to a labeled validation subset (up to 1,000 examples per benchmark) for calibration, consistent with prior unlearning evaluation methodologies.
    
\end{itemize}

\textbf{Terminology}:
    NLP benchmarks like TOFU \citep{tofu} and MUSE \citep{muse} use three subsets "retain," "forget," and "holdout" to respectively represent data to keep, remove, and evaluate on. For the WMDP \citep{wmdp} benchmark, which does not include a holdout set, we use the test set as the holdout to maintain consistency across evaluations.

\section{Results and Discussion}

We now discuss our results and directly address each of the key research questions. \\

\textbf{Q1: Can the ILL structure uncover otherwise
unnoticed information about the unlearning
process?}

\begin{figure*}[ht]
    \centering
    \includegraphics[width=1.0\textwidth]
    {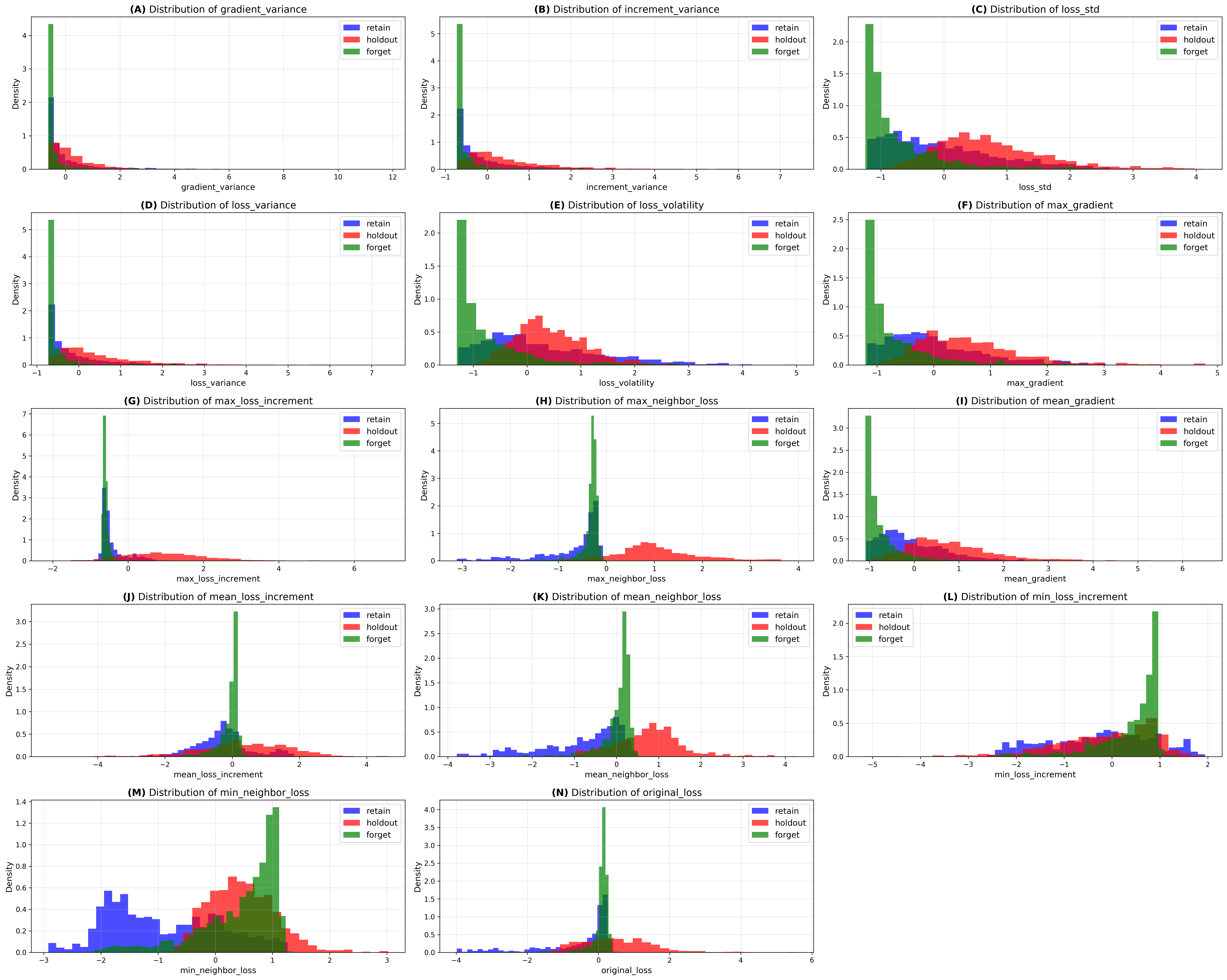}
    \caption{Histograms of 14 Input Loss Landscape (ILL) features across forgotten (green), retained (blue), and holdout (red) examples, after normalizing across all data points. These reveal structural distinctions in the local loss landscape, highlighting flattened, low-variance profiles for successfully forgotten data. 
    These plots correspond to parameters: max tokens=300, test size=0.2, neighbor percentage=0.2, number of neighbors=15.
    }
    \label{fig:feature_histograms}
\end{figure*}

Figure~\ref{fig:feature_histograms} displays the feature values distribution across retained, forgotten, and holdout examples.

The ILL around forgotten points reveals clear structural differences compared to retained or holdout points despite overlapping input sentence loss values (histogram: \textbf{N}). Forgotten samples show highly concentrated, low-variance distributions across nearly all metrics, including low loss variance between neighbors(\textbf{D}), low gradient magnitude from the input sentence to its neighbors w.r.t. embeddings (\textbf{A, F, I}), low neighbors’ loss volatility (\textbf{E}), and low loss increments from input sentences to neighbors (\textbf{B, G, J, L}), indicating a flat and shallow loss surface. In contrast, retained and holdout samples display broader, heavy-tailed distributions reflecting active learning dynamics and varied local curvatures. Specifically, mean, minimum, and maximum neighbor losses consistently separate the groups, with retained samples typically residing in sharp, well-fitted basins (\textbf{A, B, F, G, I}), holdout samples occupying more volatile or unoptimized regions (\textbf{C, D, E, G, J, L}), and forgotten samples lying in flat, low-sensitivity zones. Together, these findings suggest that unlearning reshapes the loss landscape by suppressing gradients and local curvature around forgotten samples, effectively removing the model’s sensitivity to those input regions rather than merely altering loss values.

\textbf{Q2: How does REMIND compare to pointwise and black-box baselines in uncovering forgetting patterns?}

\input{tables/aggregate_comparison.tex}

We compare \textsc{REMIND} against baselines including output difference, pointwise loss delta, and membership inference-style scores, and find that across benchmarks, \textsc{REMIND} achieves higher sensitivity and lower false-negative rates in detecting unlearning failures, demonstrating the effectiveness of our evaluation methodology (Tables~\ref{tab:aggregate_comparison_part1} and~\ref{tab:aggregate_comparison_part2}).
Unlike adapted black-box MIAs that incorporate neighborhood statistics (e.g., maximum or mean values), \textsc{REMIND} leverages full ILL geometry, enabling more efficient and robust detection. This deeper neighborhood analysis uncovers subtle residual memorization overlooked by other methods, allowing \textsc{REMIND} to successfully flag unlearning failures that baselines misclassify as successful forgetting. These findings indicate that the ILL encodes persistent memorization “footprints” of training examples that pointwise or summary-based methods fail to capture, and highlight \textsc{REMIND}'s uniquely sensitive view into the model’s residual learning state.

\textbf{Q3: Do ILL-based signals remain consistently informative across various samples, architectures, and unlearning techniques?}

We test the robustness of \textsc{REMIND} across multiple axes of variation—including paraphrased inputs, different model architectures, benchmark datasets, and unlearning algorithms—and report results in Table~\ref{tab:detailed_scores_orig}. 

\textsc{REMIND} maintains consistently high performance across all settings, while baseline scores are generally lower, especially under paraphrasing. This consistency across diverse conditions highlights \textsc{REMIND}'s robustness and supports its use as a practical, plug-in evaluation method for real-world scenarios where both inputs and model architectures may vary.

\input{tables/detailed_scores_orig.tex}

\section{Conclusions}

This work presents several key contributions to the evaluation of machine memorization and unlearning. First, we show that unlearning reshapes the geometry of the Input Loss Landscape (ILL), allowing residual memorization to be detected through structural analysis rather than solely via pointwise metrics. 
Second, we demonstrate that forgetting is not binary but instead produces intermediate, structured signatures in the loss landscape. 
Third, we introduce \textsc{REMIND}, a novel black-box evaluation method that leverages these structural patterns to detect unlearning failures missed by conventional metrics, and show that it consistently outperforms both pointwise and neighborhood-based baselines across a wide range of models and datasets.
Fourth, we propose an embedding-proximity perturbation technique that is semantically grounded, efficient, and tokenizer-free, enabling robust probing of model behavior without requiring access to internal model components.
Finally, we highlight \textsc{REMIND}'s potential for monitoring unlearning progress during training by tracking convergence in ILL feature space and point to a potential limitation and tradeoff in the unlearning process.

\textbf{ILL Geometry and Intermediate Forgetting Structure.}  
Unlearning affects not only the loss on a forgotten input but also reshapes the surrounding ILL, which can be probed through black-box evaluation of semantically similar inputs. Forgotten examples occupy an intermediate space in the ILL: they exhibit flatter, less volatile neighborhoods than retained and holdout points, but with values in between them. These findings indicate that forgetting produces smooth but not featureless regions in the loss landscape, allowing for finer-grained detection based on local curvature and neighborhood statistics.

\textbf{\textsc{REMIND} as a Novel Evaluation Method.}  
By analyzing the full geometry of the ILL using structured perturbations, \textsc{REMIND} identifies subtle memorization "footprints" left by forgotten samples—signals often missed by pointwise or summary-based neighborhood metrics. It consistently performs well across paraphrased inputs, model architectures, and unlearning methods, and relies only on black-box access, making it suitable for real-world deployments where model internals are unavailable. Empirically, \textsc{REMIND} outperforms standard baselines, pointwise methods, as well as neighborhood-based black-box MIAs, offering clearer separability between forgotten, retained, and holdout examples. These results support \textsc{REMIND} as a sensitive, practical, and deployment-ready tool for verifying unlearning effectiveness in diverse settings.

\textbf{Embedding-Proximity Perturbation Technique.}  
An additional contribution of this work is our embedding-proximity perturbation technique, which enables structured, meaning-preserving variation of inputs without relying on model-specific tokenizers or internal access. By generating neighbors through token-level embedding similarity, the method efficiently explores the local loss landscape in a semantically grounded way. It avoids the limitations of prior approaches that depend on random noise, handcrafted edits, or costly MLM queries, providing a scalable and interpretable alternative for evaluating unlearning and membership inference.

\textbf{Unlearning Monitoring and Future Directions.}  
Beyond post-hoc evaluation, \textsc{REMIND} offers potential for monitoring unlearning progress during training by tracking convergence patterns in the ILL feature space. Preliminary analysis (Figure~\ref{fig:feature_histograms}) suggests that unlearning increases the loss on forgotten samples, but this increase appears bounded by the maximum loss observed on holdout data—a potential constraint on generalization that warrants further study. These findings expose limitations in current unlearning techniques and point to a deeper tradeoff between effective forgetting and model generalization. Future work will explore these dynamics and build upon the framework described.

\section{Bibliographical References}

\bibliographystyle{lrec2026-natbib}
\bibliography{references}

\end{document}

%% file: tables/aggregate_comparison.tex
\begin{table*}[t]
    \centering
    {\setlength{\tabcolsep}{2pt}
    \scriptsize
    \rowcolors{2}{gray!15}{white}
    \begin{tabular}{|l|cc|cc|cc|cc|}
        \hline
        \textbf{Method} 
        & \multicolumn{2}{c|}{retain\_vs\_all\_auc} 
        & \multicolumn{2}{c|}{forget\_vs\_all\_auc} 
        & \multicolumn{2}{c|}{holdout\_vs\_all\_auc} 
        & \multicolumn{2}{c|}{multi\_class\_auc} \\
        \cline{2-9}
        & Orig & Reph 
        & Orig & Reph 
        & Orig & Reph 
        & Orig & Reph \\
        \hline
        Zlib Compression \citep{zlib}       & 66.1871 & 55.39 & 64.3143 & 62.0757 & 21.3857 & 30.6443 & 50.63 & 49.37 \\
        MIN-K\%++ \citep{minkpp}               & 52.8057 & 45.28 & 51.6614 & 54.8843 & 45.5343 & 49.8371 & 50 & 50 \\
        ROUGE-L F1 \citep{rouge_lin_2004}             & 63.1033 & 57.3317 & 53.8467 & 56.7017 & 66.46 & 60.6867 & 62.1733 & 58.355 \\
        Simplified SPV-MIA-mean \citep{SPV_MIA}         & 47.1329 & 46.6243 & 51.1714 & 48.0386 & 51.6914 & 55.3357 & 50 & 50 \\
        Simplified SPV-MIA-max  \citep{SPV_MIA}         & 44.4043 & 44.6557 & 55.9643 & 51.5186 & 49.6329 & 53.8229 & 50 & 50 \\
        Loss based            & 30.0914 & 40.9029 & 48.5271 & 46.9657 & 71.3814 & 62.1329 & 50 & 50 \\
        Min-k\% \citep{detecting_pretraining_data}        & 55.3386 & 48.9414 & 41.2571 & 49.34 & 53.4043 & 51.7229 & 50 & 50 \\
    
        \hline
        REMIND (ours): Random Forest    & \underline{82.1814} & \underline{72.5029} & \textbf{80.6186} & \textbf{71.6086} & \textbf{85.38} & \textbf{78.4871} & \textbf{82.84} & \underline{74.2571} \\
        REMIND (ours): Logistic Regression   & \textbf{82.3771} & \textbf{73.5471} & \underline{77.8186} & \underline{71.29} & \underline{82.6514} & \underline{77.9829} & \underline{82.2214} & \textbf{75.2329} \\
        \hline
    \end{tabular}
    }
    \caption{Aggregate Comparison of Unlearning Evaluation Metrics (Part 1: AUC Metrics)}
    \label{tab:aggregate_comparison_part1}
\end{table*}

\begin{table*}[t]
    \centering
    {\setlength{\tabcolsep}{2pt}
    \scriptsize
    \rowcolors{2}{gray!15}{white}
    \begin{tabular}{|l|cc|cc|cc|}
        \hline
        \textbf{Method} 
        & \multicolumn{2}{c|}{retain\_vs\_all\_auc\_at\_1\_fp} 
        & \multicolumn{2}{c|}{forget\_vs\_all\_auc\_at\_1\_fp} 
        & \multicolumn{2}{c|}{holdout\_vs\_all\_auc\_at\_1\_fp} \\
        \cline{2-7}
        & Orig & Reph 
        & Orig & Reph 
        & Orig & Reph \\
        \hline
        Zlib Compression \citep{zlib}       & 13.3029 & 5.28 & 1.38286 & 1.69143 & 0.217143 & 0.478571 \\
        MIN-K\%++ \citep{minkpp}               & 3.08 & 1.68 & 1.87714 & 2.90714 & 0.184286 & 0.4 \\
        ROUGE-L F1 \citep{rouge_lin_2004}              & 4.84667 & 2.405 & 1.79167 & 2.18 & 2.69667 & 2.38833 \\
        Simplified SPV-MIA-mean \citep{SPV_MIA}          & 0.718571 & 0.715714 & 8.72714 & 1.34714 & 9.26429 & 2.47714 \\
        Simplified SPV-MIA-max \citep{SPV_MIA}           & 0.5 & 0.771429 & 7.93429 & 1.26143 & 9.19143 & 2.94143 \\
        Loss based             & 1.24857 & 1.72286 & 1.04857 & 0.838571 & 24.4 & 17.8886 \\
        Min-k\% \citep{detecting_pretraining_data}         & 5.07143 & 2.25286 & 1.20429 & 1.57857 & 0.338571 & 0.582857 \\
                \hline
        REMIND (ours): Random Forest    & \textbf{28.9871} & \textbf{14.0514} & \textbf{28.6229} & \textbf{16.5729} & \textbf{49.6657} & \textbf{39.02} \\
        REMIND (ours): Logistic Regression   & \underline{26.0386} & \underline{12.0371} & \underline{21.9614} & \underline{12.7243} & \underline{43.7243} & \underline{35.8529} \\
        \hline
    \end{tabular}
    }
    \caption{Aggregate Comparison of Unlearning Evaluation Metrics (Part 2: AUC at 1\% FPR Metrics)}
    \label{tab:aggregate_comparison_part2}
\end{table*}

%% file: tables/detailed_scores_orig.tex
\begin{table*}[htb]
    \centering
    {\setlength{\tabcolsep}{2pt}
    \renewcommand\arraystretch{1.1}
    \tiny
    \rowcolors{2}{gray!15}{white}
    \begin{tabular}{|l|l|l|l|c|c|c|c|c|c|c|c|c|c|c|}
        \hline
        \textbf{Model} & \textbf{Benchmark} & \textbf{Method} & \textbf{Classifier} 
        & \makecell[c]{Retain vs\\All AUC}
        & \makecell[c]{Forget vs\\All AUC}
        & \makecell[c]{Holdout vs\\All AUC}
        & \makecell[c]{Retain vs\\All AUC@1FP}
        & \makecell[c]{Forget vs\\All AUC@1FP}
        & \makecell[c]{Holdout vs\\All AUC@1FP}
        & \makecell[c]{Retained vs\\Forgotten}
        & \makecell[c]{Forgotten vs\\Holdout}
        & \makecell[c]{Overall\\Score}
        & \makecell[c]{Accuracy}
        & \makecell[c]{F1} \\
        \hline
        llama-3-8b & WMDP & ELM & LogReg & 99.564 & 98.464 & 100 & 95.455 & 93.434 & 100 & 99.64 & 100 & 100 & 100 & 100 \\
        llama-3-8b & WMDP & ELM & Tree & 99.688 & 99.85 & 100 & 95.96 & 94.949 & 100 & 99.768 & 100 & 100 & 100 & 100 \\
        llama-3-8b & WMDP & TAR & LogReg & 92.596 & 79.734 & 74.829 & 39.394 & 12.626 & 5.051 & 84.098 & 67.899 & 41.231 & 67.899 & 41.231 \\
        llama-3-8b & WMDP & TAR & Tree & 94.341 & 87.947 & 95.388 & 45.914 & 32.798 & 34.833 & 91.77 & 87.899 & 81.053 & 87.899 & 81.053 \\
        llama-3-8b & WMDP & PBJ & LogReg & 92.843 & 77.174 & 88.969 & 24.121 & 17.588 & 28.643 & 86.132 & 79.195 & 66.304 & 79.195 & 66.304 \\
        llama-3-8b & WMDP & PBJ & Tree & 93.62 & 86.26 & 96.825 & 46.704 & 27.09 & 43.445 & 92.169 & 90.604 & 86.603 & 90.604 & 86.603 \\
        llama-3-8b & WMDP & RMU-LAT & LogReg & 95.29 & 94.284 & 99.959 & 17.172 & 38.889 & 99.495 & 96.311 & 99.664 & 99.492 & 99.664 & 99.492 \\
        llama-3-8b & WMDP & RMU-LAT & Tree & 96.661 & 96.189 & 99.685 & 37.185 & 58.311 & 99.495 & 96.402 & 99.664 & 99.495 & 99.664 & 99.495 \\
        llama-3-8b & WMDP & RMU & LogReg & 94.614 & 94.787 & 99.611 & 46.465 & 26.263 & 98.485 & 95.624 & 98.655 & 97.949 & 98.655 & 97.949 \\
        llama-3-8b & WMDP & RMU & Tree & 94.867 & 94.374 & 99.968 & 51.439 & 23.677 & 98.99 & 95.235 & 99.328 & 98.985 & 99.328 & 98.985 \\
        zephyr-7b & WMDP & SimNPO & LogReg & 82.424 & 83.889 & 99.893 & 2.525 & 7.071 & 95.96 & 87.851 & 97.815 & 96.692 & 97.815 & 96.692 \\
        zephyr-7b & WMDP & SimNPO & Tree & 82.746 & 84.675 & 99.944 & 6.053 & 12.106 & 94.949 & 87.807 & 98.151 & 97.201 & 98.151 & 97.201 \\
        llama-2-7b & TOFU-forget05 & SimNPO & LogReg & 60.703 & 57.508 & 57.211 & 0 & 0 & 0 & 60.31 & 66.25 & 4.706 & 66.25 & 4.706 \\
        llama-2-7b & TOFU-forget05 & SimNPO & Tree & 53.848 & 54.301 & 51.688 & 0 & 7.75 & 1.25 & 55.845 & 62.5 & 25 & 62.5 & 25 \\
        llama-2-7b & TOFU-forget10 & SimNPO & LogReg & 72.883 & 79.117 & 58.969 & 5 & 27.5 & 0 & 73.964 & 62.5 & 8.163 & 62.5 & 8.163 \\
        llama-2-7b & TOFU-forget10 & SimNPO & Tree & 71.969 & 75.27 & 64.371 & 0 & 24.75 & 0 & 72.241 & 66.667 & 37.5 & 66.667 & 37.5 \\
        llama-2-7b & MUSE-News & SimNPO & LogReg & 88.923 & 70.741 & 97.327 & 19.774 & 1.13 & 46.893 & 90.618 & 92.279 & 88.319 & 92.279 & 88.319 \\
        llama-2-7b & MUSE-News & SimNPO & Tree & 88.705 & 82.847 & 97.245 & 17.407 & 11.345 & 39.058 & 91.489 & 92.09 & 87.719 & 92.09 & 87.719 \\
        llama-2-7b & MUSE-Books & SimNPO & LogReg & 50.799 & 59.375 & 58.981 & 0 & 1.695 & 1.13 & 55.36 & 66.729 & 0 & 66.729 & 0 \\
        llama-2-7b & MUSE-Books & SimNPO & Tree & 53.475 & 56.534 & 56.09 & 4.675 & 0.876 & 0.565 & 55.685 & 65.789 & 25.41 & 65.789 & 25.41 \\
        \hline
    \end{tabular}
    }
    \caption{Detailed scores for individual models - Original Inputs. The results are based on the following parameters: m=20, K\_neighbors=15, replacement probability $p$=0.3, test size=0.2. For clarity in this large table, we have not inserted citations; the citations can be found in the related work and the experimental setup}
    \label{tab:detailed_scores_orig}
\end{table*}

%% file: references.bib
@article{machine_unlearning_survey,
  title={A Comprehensive Survey of Machine Unlearning Techniques for Large Language Models},
  author={Geng, Jiahui and Li, Qing and Woisetschlaeger, Herbert and Chen, Zongxiong and Cai, Fengyu and Wang, Yuxia and Nakov, Preslav and Jacobsen, Hans-Arno and Karray, Fakhri},
  journal={arXiv preprint arXiv:2503.01854},
  year={2025},
  url={https://arxiv.org/abs/2503.01854}
}

@article{critical_reexamination,
  title   = {Position: LLM Unlearning Benchmarks are Weak Measures of Progress},
  author  = {Pratiksha Thaker and Shengyuan Hu and Neil Kale and Yash Maurya and Zhiwei Steven Wu and Virginia Smith},
  journal = {arXiv preprint arXiv:2410.02879},
  year    = {2024},
  url     = {https://arxiv.org/abs/2410.02879}
}

@article{recall,
  title={ReCaLL: Membership Inference via Relative Conditional Log-Likelihoods},
  author={Xie, Roy and Wang, Junlin and Huang, Ruomin and Zhang, Minxing and Ge, Rong and Pei, Jian and Gong, Neil Zhenqiang and Dhingra, Bhuwan},
  journal={arXiv preprint arXiv:2406.15968},
  year={2024},
  url={https://arxiv.org/abs/2406.15968}
}

@article{minkpp,
  title={Min-K\%++: Improved Baseline for Detecting Pre-Training Data from Large Language Models},
  author={Zhang, Jingyang and Sun, Jingwei and Yeats, Eric and Ouyang, Yang and Kuo, Martin and Zhang, Jianyi and Yang, Hao Frank and Li, Hai},
  journal={arXiv preprint arXiv:2404.02936},
  year={2024},
  url={https://arxiv.org/abs/2404.02936}
}

@article{detecting_pretraining_data,
  title={Detecting Pretraining Data from Large Language Models},
  author={Shi, Weijia and Ajith, Anirudh and Xia, Mengzhou and Huang, Yangsibo and Liu, Daogao and Blevins, Terra and Chen, Danqi and Zettlemoyer, Luke},
  journal={arXiv preprint arXiv:2310.16789},
  year={2023},
  url={https://arxiv.org/abs/2310.16789}
}

@article{tofu,
  title={TOFU: A Task of Fictitious Unlearning for LLMs},
  author={Maini, Pratyush and Feng, Zhili and Schwarzschild, Avi and Lipton, Zachary C. and Kolter, J. Zico},
  journal={arXiv preprint arXiv:2401.06121},
  year={2024},
  url={https://arxiv.org/abs/2401.06121}
}

@article{muse,
  title={MUSE: Machine Unlearning Six-Way Evaluation for Language Models},
  author={Shi, Weijia and Lee, Jaechan and Huang, Yangsibo and Malladi, Sadhika and Zhao, Jieyu and Holtzman, Ari and Liu, Daogao and Zettlemoyer, Luke and Smith, Noah A. and Zhang, Chiyuan},
  journal={arXiv preprint arXiv:2407.06460},
  year={2024},
  url={https://arxiv.org/abs/2407.06460}
}

@article{wmdp,
  title={The WMDP Benchmark: Measuring and Reducing Malicious Use With Unlearning},
  author={Li, Nathaniel and Pan, Alexander and Gopal, Anjali and Yue, Summer and Berrios, Daniel and Gatti, Alice and Li, Justin D. and Dombrowski, Ann-Kathrin and Goel, Shashwat and Phan, Long and Mukobi, Gabriel and Helm-Burger, Nathan and Lababidi, Rassin and Justen, Lennart and Liu, Andrew B. and Chen, Michael and Barrass, Isabelle and Zhang, Oliver and Zhu, Xiaoyuan and Tamirisa, Rishub and Bharathi, Bhrugu and Khoja, Adam and Zhao, Zhenqi and Herbert-Voss, Ariel and Breuer, Cort B. and Marks, Samuel and Patel, Oam and Zou, Andy and Mazeika, Mantas and Wang, Zifan and Oswal, Palash and Lin, Weiran and Hunt, Adam A. and Tienken-Harder, Justin and Shih, Kevin Y. and Talley, Kemper and Guan, John and Kaplan, Russell and Steneker, Ian and Campbell, David and Jokubaitis, Brad and Levinson, Alex and Wang, Jean and Qian, William and Karmakar, Kallol Krishna and Basart, Steven and Fitz, Stephen and Levine, Mindy and Kumaraguru, Ponnurangam and Tupakula, Uday and Varadharajan, Vijay and Wang, Ruoyu and Shoshitaishvili, Yan and Ba, Jimmy and Esvelt, Kevin M. and Wang, Alexandr and Hendrycks, Dan},
  journal={arXiv preprint arXiv:2403.03218},
  year={2024},
  url={https://arxiv.org/abs/2403.03218}
}

@article{rwku,
  title={RWKU: Benchmarking Real-World Knowledge Unlearning for Large Language Models},
  author={Jin, Zhuoran and Cao, Pengfei and Wang, Chenhao and He, Zhitao and Yuan, Hongbang and Li, Jiachun and Chen, Yubo and Liu, Kang and Zhao, Jun},
  journal={arXiv preprint arXiv:2406.10890},
  year={2024},
  url={https://arxiv.org/abs/2406.10890}
}

@article{inexact_unlearning,
  title={Inexact Unlearning Needs More Careful Evaluations to Avoid a False Sense of Privacy},
  author={Hayes, Jamie and Shumailov, Ilia and Triantafillou, Eleni and Khalifa, Amr and Papernot, Nicolas},
  journal={arXiv preprint arXiv:2403.01218},
  year={2024},
  url={https://arxiv.org/abs/2403.01218}
}

@article{openunlearning,
  title={OpenUnlearning: Accelerating LLM Unlearning via Unified Benchmarking of Methods and Metrics},
  author={Dorna, Vineeth and Mekala, Anmol and Zhao, Wenlong and McCallum, Andrew and Lipton, Zachary C. and Kolter, J. Zico and Maini, Pratyush},
  journal={arXiv preprint arXiv:2506.12618},
  year={2024},
  url={https://arxiv.org/abs/2506.12618}
}

@article{adversarial_machine_learning,
  title={Adversarial Attacks on Large Language Models Using Regularized Relaxation},
  author={Chacko, Samuel Jacob and Biswas, Sajib and Islam, Chashi Mahiul and Liza, Fatema Tabassum and Liu, Xiuwen},
  journal={arXiv preprint arXiv:2410.19160},
  year={2024},
  url={https://arxiv.org/abs/2410.19160}
}

@misc{gdpr,
  author = {{Council of European Union}},
  title = {{Council Regulation (EU) No 2016/679}},
  year = {2016},
  note = {General Data Protection Regulation},
  url = {https://eur-lex.europa.eu/eli/reg/2016/679/oj}
}

@article{unlearningTaxonomy,
  title={Machine Unlearning: Taxonomy, Metrics, Applications, Challenges, and Prospects},
  author={Li, Na and Zhou, Chunyi and Gao, Yansong and Chen, Hui and Fu, Anmin and Zhang, Zhi and Yu, Shui},
  journal={arXiv preprint arXiv:2403.08254},
  year={2024},
  url={https://arxiv.org/pdf/2403.08254}
}

@inproceedings{rouge_lin_2004,
  title={ROUGE: A Package for Automatic Evaluation of Summaries},
  author={Lin, Chin-Yew},
  booktitle={Text Summarization Branches Out: Proceedings of the ACL-04 Workshop},
  pages={74--81},
  year={2004},
  address={Barcelona, Spain},
  publisher={Association for Computational Linguistics},
  url={https://aclanthology.org/W04-1013.pdf}
}

@inproceedings{zlib,
  title={Extracting Training Data from Large Language Models},
  author={Carlini, Nicholas and Tram{\`e}r, Florian and Wallace, Eric and Jagielski, Matthew and Herbert-Voss, Ariel and Lee, Katherine and Roberts, Adam and Brown, Tom and Song, Dawn and Erlingsson, {\'U}lfar and Oprea, Alina and Raffel, Colin},
  booktitle={30th USENIX Security Symposium (USENIX Security 21)},
  year={2021},
  organization={USENIX Association},
  url={https://www.usenix.org/system/files/sec21-carlini-extracting.pdf}
}

@article{intrinsic_parametric_knowledge_traces,
  title={Intrinsic Evaluation of Unlearning Using Parametric Knowledge Traces},
  author={Hong, Yihuai and Yu, Lei and Yang, Haiqin and Ravfogel, Shauli and Geva, Mor},
  journal={arXiv preprint arXiv:2406.11614},
  year={2025},
  url={https://arxiv.org/pdf/2406.11614}
}

@article{touvron2023llama2,
  title={Llama 2: Open Foundation and Fine-Tuned Chat Models},
  author={Touvron, Hugo and Martin, Louis and Stone, Kevin and Albert, Peter and Almahairi, Yasmine and Babaei, Nikolay and Bashlykov, Nikolay and Batra, Soumya and Bhargava, Prajjwal and Bhosale, Shruti and Bikel, Dan and Blecher, Lukas and Canton Ferrer, Cristian and Chen, Moya and Cucurull, Guillem and Esiobu, David and Fernandes, Jude and Fu, Jeremy and Fu, Wenyin and Fuller, Brian and Gao, Cynthia and Goswami, Vedanuj and Goyal, Naman and Hartshorn, Anthony and Hosseini, Saghar and Hou, Rui and Inan, Hakan and Kardas, Marcin and Kerkez, Viktor and Khabsa, Madian and Kloumann, Isabel and Korenev, Artem and Koura, Punit Singh and Lachaux, Marie-Anne and Lavril, Thibaut and Lee, Jenya and Liskovich, Diana and Lu, Yinghai and Mao, Yuning and Martinet, Xavier and Mihaylov, Todor and Mishra, Pushkar and Molybog, Igor and Nie, Yixin and Poulton, Andrew and Reizenstein, Jeremy and Rungta, Rashi and Saladi, Kalyan and Schelten, Alan and Silva, Ruan and Smith, Eric Michael and Subramanian, Ranjan and Tan, Xiaoqing Ellen and Tang, Binh and Taylor, Ross and Williams, Adina and Xiang Kuan, Jian and Xu, Puxin and Yan, Zheng and Zarov, Iliyan and Zhang, Yuchen and Fan, Angela and Kambadur, Melanie and Narang, Sharan and Rodriguez, Aurelien and Stojnic, Robert and Edunov, Sergey and Scialom, Thomas},
  journal={arXiv preprint arXiv:2307.09288},
  year={2023},
  url={https://arxiv.org/pdf/2307.09288.pdf}
}

@article{grattafiori2024llama3,
  title={Llama 3: The Reference Implementation},
  author={Grattafiori, Arthur and Dubey, Abhimanyu and Jauhri, Abhinav and Pandey, Abhishek and Kadian, Abhishek and Al-Dahle, Ahmad and Letman, Anton and Mathur, Aran and Schelten, Beren and Vaughan, Blake and others},
  journal={arXiv preprint arXiv:2407.21783},
  year={2024},
  url={https://arxiv.org/pdf/2407.21783.pdf}
}

@article{tunstall2023zephyr,
  title={Zephyr: Direct Distillation of LM Alignment},
  author={Tunstall, Lewis and Beeching, Edward and Lambert, Nathan and Rajani, Nazneen and Rasul, Kashif and Belkada, Younes and Huang, Shengyi and von Werra, Leandro and Fourrier, Clémentine and Habib, Nathan and others},
  journal={arXiv preprint arXiv:2310.16944},
  year={2023},
  url={https://arxiv.org/pdf/2310.16944.pdf}
}

@misc{remove,
  title={REGULATION (EU) 2016/679 OF THE EUROPEAN PARLIAMENT AND OF THE COUNCIL of 27 April 2016 on the protection of natural persons with regard to the processing of personal data and on the free movement of such data, and repealing Directive 95/46/EC},
  author={European Parliament and Council},
  year={2016},
  url={https://eur-lex.europa.eu/legal-content/EN/TXT/?uri=CELEX:32016R0679}
}

@inproceedings{hsu2025are,
  title={Are We Really Unlearning? The Presence of Residual Knowledge in Machine Unlearning},
  author={Hsiang Hsu and Pradeep Niroula and Zichang He and Chun-Fu Chen},
  booktitle={I Can't Believe It's Not Better: Challenges in Applied Deep Learning},
  year={2025},
  url={https://openreview.net/forum?id=HsjHGNYv2O}
}

@article{NoisyNeighbors,
  title={Noisy Neighbors: Efficient membership inference attacks against LLMs},
  author={Galli, Filippo and Melis, Luca and Cucinotta, Tommaso},
  journal={arXiv preprint arXiv:2406.16565},
  year={2024}
}

@inproceedings{SPV_MIA,
  author = {Fu, Wenjie and Wang, Huandong and Gao, Chen and Liu, Guanghua and Li, Yong and Jiang, Tao},
  booktitle = {Advances in Neural Information Processing Systems},
  editor = {A. Globerson and L. Mackey and D. Belgrave and A. Fan and U. Paquet and J. Tomczak and C. Zhang},
  pages = {134981--135010},
  publisher = {Curran Associates, Inc.},
  title = {Membership Inference Attacks against Fine-tuned Large Language Models via Self-prompt Calibration},
  url = {https://proceedings.neurips.cc/paper_files/paper/2024/file/f36ad694188bb4c4bbbd61e2038e069e-Paper-Conference.pdf},
  volume = {37},
  year = {2024}
}

@article{neighbourhood_comparison,
  title={Membership inference attacks against language models via neighbourhood comparison},
  author={Mattern, Justus and Mireshghallah, Fatemehsadat and Jin, Zhijing and Sch{\"o}lkopf, Bernhard and Sachan, Mrinmaya and Berg-Kirkpatrick, Taylor},
  journal={arXiv preprint arXiv:2305.18462},
  year={2023}
}

@article{SMIA,
  title={Semantic membership inference attack against large language models},
  author={Mozaffari, Hamid and Marathe, Virendra J},
  journal={arXiv preprint arXiv:2406.10218},
  year={2024}
}

@article{xu2024targeted,
  title={Targeted training data extraction—neighborhood comparison-based membership inference attacks in large language models},
  author={Xu, Huan and Zhang, Zhanhao and Yu, Xiaodong and Wu, Yingbo and Zha, Zhiyong and Xu, Bo and Xu, Wenfeng and Hu, Menglan and Peng, Kai},
  journal={Applied Sciences},
  volume={14},
  number={16},
  pages={7118},
  year={2024}
}

@article{ILL_robust_overfitting,
  title={Understanding and combating robust overfitting via input loss landscape analysis and regularization},
  author={Li, Lin and Spratling, Michael},
  journal={Pattern Recognition},
  volume={136},
  pages={109229},
  year={2023},
  publisher={Elsevier}
}

@article{ILL_adversarial_weight_perturbation,
  title={Adversarial weight perturbation helps robust generalization},
  author={Wu, Dongxian and Xia, Shu-Tao and Wang, Yisen},
  journal={Advances in Neural Information Processing Systems},
  volume={33},
  pages={2958--2969},
  year={2020}
}

@inproceedings{ILL_sharpness_adversarial_samples,
  title={Detecting Adversarial Samples through Sharpness of Loss Landscape},
  author={Zheng, Rui and Dou, Shihan and Zhou, Yuhao and Liu, Qin and Gui, Tao and Zhang, Qi and Wei, Zhongyu and Huang, Xuan-Jing and Zhang, Menghan},
  booktitle={Findings of the Association for Computational Linguistics: ACL 2023},
  pages={11282--11298},
  year={2023}
}

@inproceedings{ILL_input_gradient_regularization,
  title={Improving the adversarial robustness and interpretability of deep neural networks by regularizing their input gradients},
  author={Ross, Andrew and Doshi-Velez, Finale},
  booktitle={Proceedings of the AAAI Conference on Artificial Intelligence},
  volume={32},
  year={2018}
}

@article{ELM,
  title={Erasing conceptual knowledge from language models},
  author={Gandikota, Rohit and Feucht, Sheridan and Marks, Samuel and Bau, David},
  journal={arXiv preprint arXiv:2410.02760},
  year={2024}
}

@article{TAR,
  title={Tamper-resistant safeguards for open-weight llms},
  author={Tamirisa, Rishub and Bharathi, Bhrugu and Phan, Long and Zhou, Andy and Gatti, Alice and Suresh, Tarun and Lin, Maxwell and Wang, Justin and Wang, Rowan and Arel, Ron and others},
  journal={arXiv preprint arXiv:2408.00761},
  year={2024}
}

@article{RMU,
  title={The wmdp benchmark: Measuring and reducing malicious use with unlearning},
  author={Li, Nathaniel and Pan, Alexander and Gopal, Anjali and Yue, Summer and Berrios, Daniel and Gatti, Alice and Li, Justin D and Dombrowski, Ann-Kathrin and Goel, Shashwat and Phan, Long and others},
  journal={arXiv preprint arXiv:2403.03218},
  year={2024}
}

@article{simNPO,
  title={Simplicity prevails: Rethinking negative preference optimization for llm unlearning},
  author={Fan, Chongyu and Liu, Jiancheng and Lin, Licong and Jia, Jinghan and Zhang, Ruiqi and Mei, Song and Liu, Sijia},
  journal={arXiv preprint arXiv:2410.07163},
  year={2024}
}

@article{tokens_swapping,
  title={Data contamination calibration for black-box llms},
  author={Ye, Wentao and Hu, Jiaqi and Li, Liyao and Wang, Haobo and Chen, Gang and Zhao, Junbo},
  journal={arXiv preprint arXiv:2405.11930},
  year={2024}
}

@article{suffix_addition,
  title={Asetf: A novel method for jailbreak attack on llms through translate suffix embeddings},
  author={Wang, Hao and Li, Hao and Huang, Minlie and Sha, Lei},
  journal={arXiv preprint arXiv:2402.16006},
  year={2024}
}

@article{GPT2,
  title={Language models are unsupervised multitask learners},
  author={Radford, Alec and Wu, Jeffrey and Child, Rewon and Luan, David and Amodei, Dario and Sutskever, Ilya and others},
  journal={OpenAI blog},
  volume={1},
  number={8},
  pages={9},
  year={2019}
}
